\documentclass{article}

\usepackage{spconf,amsmath,graphicx,hyperref,cleveref}
\usepackage{amsmath,amssymb}
\usepackage{amsthm}
\usepackage{textcomp}
\usepackage{multirow}
\usepackage{booktabs}

\title{SynergyWarpNet: Attention-Guided Cooperative Warping for Neural Portrait Animation}
\name{Shihang Li$^{1,2,3}$\qquad Zhiqiang Gong$^{2,3}$\qquad  Minming Ye$^{1}$\qquad Yue Gao$^{1,*}$\qquad Wen Yao$^{2,3,*}$\thanks{$*$ Corresponding Authors.}\thanks{$†$ This work was partly supported by the Young Elite Scientist Sponsorship Program By CAST (Grant No. YESS20240697) and the National Natural Science Foundation of China (Grant No. 92371206).}}
\address{$^{1}$ MoE Key Lab of Artificial Intelligence, AI Institute, Shanghai Jiao Tong University\\
         $^{2}$ Defense Innovation Institute, Academy of Military Science
         $^{3}$ Intelligent Game and Decision Laboratory}

\begin{document}
\ninept
\maketitle
\begin{abstract}
Recent advances in neural portrait animation have demonstrated remarked potential for applications in virtual avatars, telepresence, and digital content creation. However, traditional explicit warping approaches often struggle with accurate motion transfer or recovering missing regions, while recent attention-based warping methods, though effective, frequently suffer from high complexity and weak geometric grounding. To address these issues, we propose SynergyWarpNet, an attention-guided cooperative warping framework designed for high-fidelity talking head synthesis. Given a source portrait, a driving image, and a set of reference images, our model progressively refines the animation in three stages. First, an explicit warping module performs coarse spatial alignment between the source and driving image using 3D dense optical flow. Next, a reference-augmented correction module leverages cross-attention across 3D keypoints and texture features from multiple reference images to semantically complete occluded or distorted regions. Finally, a confidence-guided fusion module integrates the warped outputs with spatially-adaptive fusing, using a learned confidence map to balance structural alignment and visual consistency. Comprehensive evaluations on benchmark datasets demonstrate state-of-the-art performance. 
\end{abstract}
%
\section{Introduction}
\label{sec:intro}
Talking head generation \cite{guo2024liveportrait,shen2025long}, as a crucial subfield of image animation, focus on synthesizing realistic facial images by transferring motion patterns from a driving video to a source avatar while maintaining the identity information of the source subject. This rapidly advancing technology has gained substantial attention in the era of AI-generated content (AIGC), demonstrating significant potential across various applications, including digital human creation, virtual conferencing, and personalized avatar animation.

Recent progress in video-driven portrait animation has demonstrated the importance of modeling both precise spatial motion and semantic correspondence. Traditional approaches typically focus on one of two complementary modeling capabilities: explicit warping, which manipulates the source image based on explicit motion priors such as 3DMM parameters \cite{ren2021pirenderer}, latent codes \cite{wang2022latent}, or keypoints \cite{siarohin2019first, wang2021one, hong2022depth,  hong2023implicit, zhang2023metaportrait, guo2024liveportrait, zhao2025synergizing, ma2025playmate}, and attention-based warping, which leverages deep representations and attention mechanisms to establish global correspondences and hallucinate missing content \cite{mallya2022implicit, wei2024aniportrait,xie2024x,ma2024follow, yang2025megactor, gao2025faceshot}. 

Explicit warping excels at identity preservation and efficient inference, as it relies on explicitly estimated motion fields to directly deform source pixels. FOMM~\cite{siarohin2019first} utilized local affine transformation to infer dense optical flow from sparse 2D implicit keypoints, while Face Vid2vid~\cite{wang2021one} extended this to 3D implicit keypoints for free-view synthesis and more subtle motion transfer. LivePortrait~\cite{guo2024liveportrait} enhanced Face Vid2vid with improved network architecture and larger-scale training, enabling stitching and retargeting control. AppMotionComp~\cite{zhao2025synergizing} jointly learned motion and appearance codebooks with transformer-based multi-scale compensation to refine facial motion and appearance for high-fidelity talking-head generation. 
However, its single-image driving heavily depends on source completeness, causing artifacts under large pose differences. 
In contrast, attention-based warping offers distinct advantages in handling challenging scenarios where explicit warping struggle. Recent studies~\cite{wei2024aniportrait,xie2024x,ma2024follow} employed LDMs with two specialized networks: a reference network $\mathcal{R}$ for multi-scale feature extraction and a denoising network $\mathcal{D}$ for motion signal injection (e.g., pose/facial landmarks) via cross-attention. Although these methods have demonstrated remarkable animated capabilities, they often suffers from slow convergence and high computational overhead due to its reliance on high-dimensional features and global attention, making it less efficient than explicit warping. 

Rather than treating these paradigms as mutually exclusive, we argue that they encode orthogonal and synergistic capabilities: spatial precision versus semantic adaptability. To fully exploit their complementarity, we propose SynergyWarpNet, an attention-guided cooperative warping framework that tightly integrates explicit geometric warping with implicit feature refinement. Specifically, our framework comprises three meticulously designed components: (1) The Dense Optical Flow Warping (DOFW) module employs dense motion flow derived from 3D keypoints to model a coarse motion between source and driving image. (2) The Reference-Augmented Correction (RAC) module refines motion understanding by aggregating multiple reference features through hierarchical attention, enabling the model to establish robust semantic correspondences under pose and occlusion variations. (3) The Confidence-Guided Fusion (CGF) module is designed to dynamically assess the reliability of reference-guided features from RAC by a learnable confidence mask. It selectively integrates these features with DOFW's warped results through a gating mechanism, producing refined inputs for the decoder to generate the final driving result. In summary, our contributions are three-folds:

\begin{figure*}[htbp]
\centering
\includegraphics[width=0.94\linewidth]{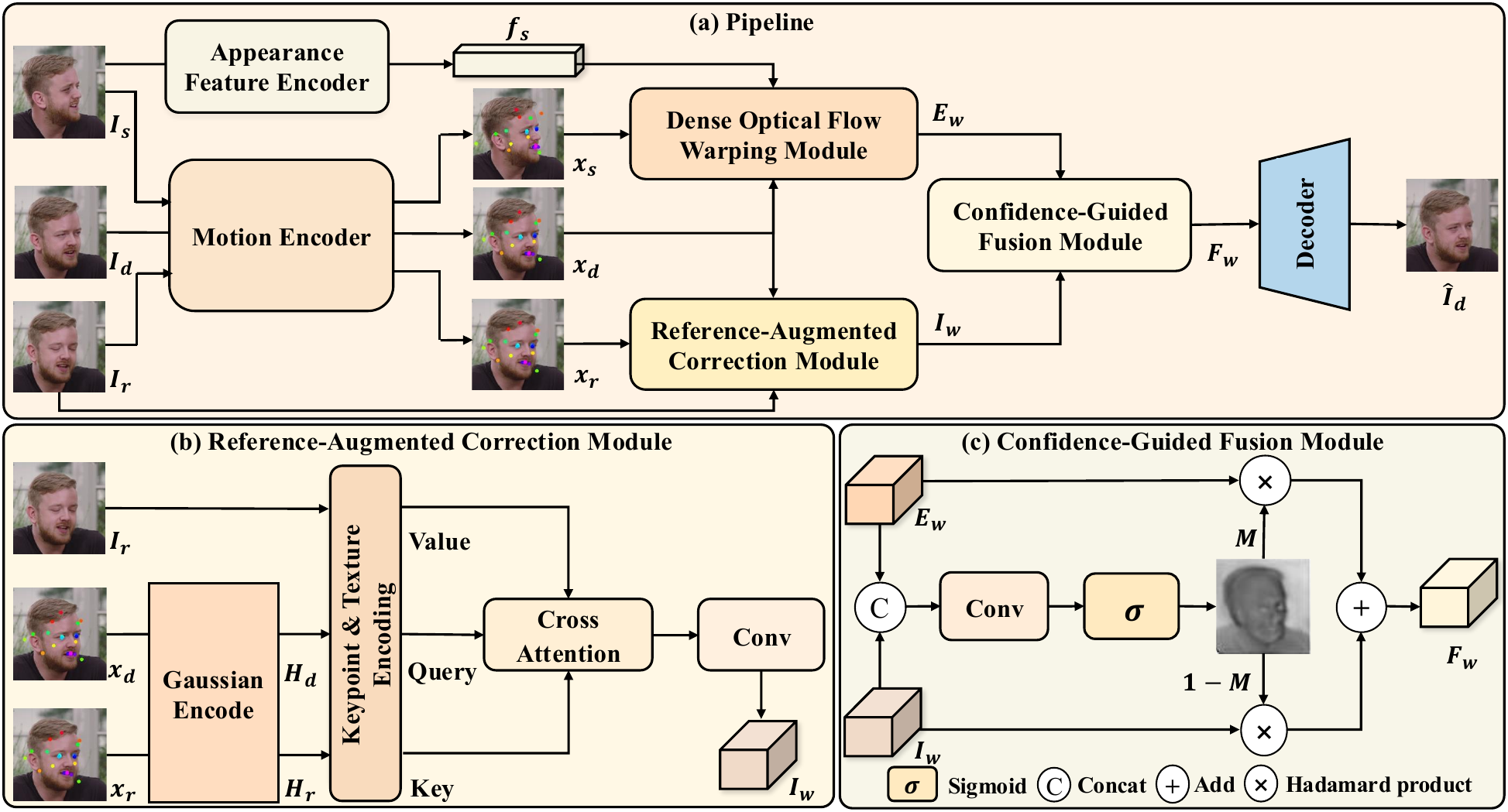}
   \caption{\textbf{Overview of our model}. Starting from a source image $I_s$, driving image $I_d$, and reference set $\{I_r^i\}^M_{i=1}$, the model first encodes appearance and motion to estimate 3D implicit keypoints. The dense optical flow warping module generates dense optical flow to align $I_s$ with $I_d$, while a reference-augmented correction module corrects occlusions and pose artifacts using multi-reference features. A confidence-guided fusion module combines both streams to produce high-fidelity output.
   }
\label{pipeline}
\vspace{-12pt}
\end{figure*}

\begin{itemize}
    \item We introduce SynergyWarpNet, an innovative unified framework that combines explicit geometry-constrained motion estimation and attention-guided region refinement in a cooperative warping architecture.
    \item To address the challenges of handling background and occlusion in explicit warping, we design a novel Reference-Augmented Correction (RAC) module to aggregate multiple reference images and a Confidence-Guided Fusion (CGF) module to fuse two warping results, enhancing motion transfer accuracy and improving information compensation.
    \item Extensive experiments demonstrate that our method achieves state-of-the-art performance across multiple benchmarks, generating high-fidelity talking head videos with superior visual quality compared to existing approaches.
\end{itemize}
\section{Method}
\label{sec:Method}
Given a single source portrait image $I_s$, driving video $\{I_d^i\}^N_{i=1}$ and a set of reference images $\{I_r^i\}^M_{i=1}$, our method aims to generate a high-fidelity talking head video $\{\hat{I}^{i}_d\}^N_{i=1}$ that accurately reproduce the facial expression and head pose specified by the driving image while maintaining source identity characteristics. \Cref{pipeline} illustrates our cascaded processing framework.
\subsection{Appearance and Motion Encoding}
\label{AME}
In this component, we employ an Appearance Encoder $\mathcal{E}$ and a Motion Encoder $\mathcal{M}$~\cite{guo2024liveportrait} to extract latent representations for appearance and motion from all input frames. The appearance feature encoder $\mathcal{E}$ transforms $I_s$ into a volumetric representation $f_s$ through depth-aware convolutions. The motion encoder $\mathcal{M}$ extracts motion information from $I_s$, $I_d$ and $I_r$, predicting a set of $\mathcal{K}$ 3D canonical keypoints $x_{c,i}$, along with the head pose $(R_i,t_i)$, expression deformation $\delta_i$ and scale factor $s_i$. Then we establish 3D keypoint correspondences $x_s, x_d$ and $x_r$ through a unified transformation $\mathcal{T}$ defined as:
\begin{equation}
x_i = \mathcal{T}(x_{c,s}, R_i, t_i, \delta_i, s_i) = s_i \cdot (x_{c,s} R_i + \delta_i) + t_i
\end{equation}
Note that both transformations share the same canonical keypoints $x_{c,s}$, ensuring the synthesized face maintains the source identity characteristics throughout the animation sequence. 

\begin{figure*}[htbp]
\centering
\includegraphics[width=1\linewidth]{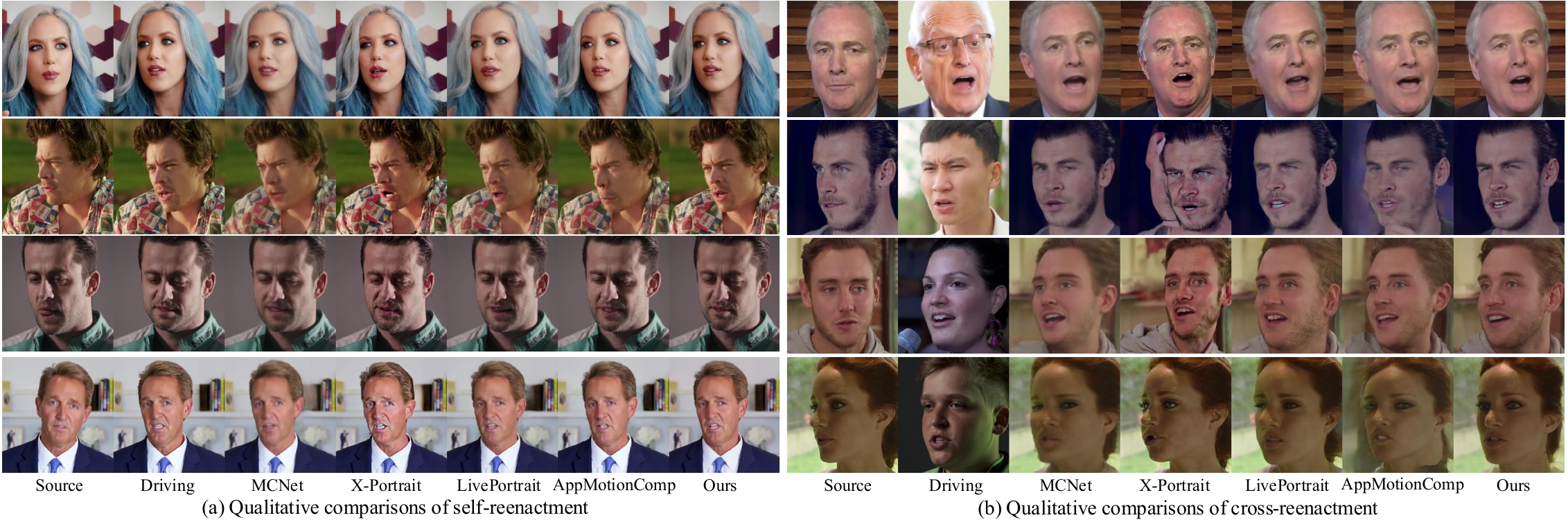}
   \caption{\textbf{Qualitative comparison with state-of-the-art methods}. (a) self-reenactment and (b) cross-reenactment on the VFHQ and HDTF datasets. Our method achieves higher motion transfer accuracy while preserving the identity.
   }
\label{qualitative}
\vspace{-10pt}
\end{figure*}

\subsection{Dense Optical Flow Warping Module}
We construct the geometry-driven deformation field by leveraging the 3D implicit keypoints. Specifically, we employ a Warping Module $\mathcal{W}$ which leverages the first-order approximation~\cite{siarohin2019first} to estimate a 3D optical flow field $w$ using paired 3D keypoints $\{x_s, x_d\}$. This flow field is then applied to $f_s$ through an affine transformation $\mathcal{A}$, yielding the explicit warped feature output:
\begin{equation}
E_w = \mathcal{A}(w, f_s)
\end{equation}

While this mechanism enables identity-preserving and interpretable alignment, it may introduce artifacts in regions where source content is occluded or insufficient—particularly under extreme poses. In the following stage, we address these limitations using reference-augmented correction and adaptive fusion.

\subsection{Reference-Augmented Correction Module}
In this stage, we propose a cross-modal attention module that leverages 3D implicit keypoints and appearance features as complementary cues to enhance source-driven synthesis fidelity and recover spatially coherent background details.

\noindent \textbf{3D Keypoints \& Texture Encoding}. We utilize precomputed 3D implicit keypoints $x_d$ and $x_r$, representing the structural information of both the driving and reference images. To overcome the structural representation limitation caused by keypoint sparsity, we employ Gaussian encoding scheme~\cite{wang2021one} that projects sparse keypoints into gaussian heatmaps. Formally, given a set of $K$ 3D keypoints $x_i\in \mathbb{R}^{(k,3)}$ of input image, we generate a 4D tensor $H_i$ through:
\begin{equation}
H_{i,k} = \exp\left( - \frac{\| G - x_{i,k} \|^2}{2 \sigma^2} \right)
\label{gaussian}
\end{equation}
where $H_{i,k}$ is the $k$-th channel of $H_i$, $G$ is the coordinate grid, $x_{i,k}$ denotes the coordinate of the $k$-th keypoint, and $\sigma^2$ denotes the predefined variance.

Subsequently, we employ a keypoint encoder $\mathcal{E}_{kp}$ to encode $H_d$ and $H^{1:M}_r$ into latent representation $\tilde{H}_d$ and $\tilde{H}^{1:M}_r$. Concurrently, a texture encoder $\mathcal{E}_{tex}$ processes the reference images $I^{1:M}_r$ into $\tilde{I}^{1:M}_r$.

\noindent \textbf{Cross-Modal Attention Texture Sampling}. The core of this component is a single scaled dot-product attention operation~\cite{vaswani2017attention}.
Specifically, as shown in \Cref{pipeline} (b), we employ $\tilde{H}_d$ as the query $Q$, while the reference information, consisting of the spatial representation of 3D keypoints $\tilde{H}^{1:M}_r$ and textures $\tilde{I}^{1:M}_r$, serve as the keys $K$ and values $V$, respectively. The softmax-normalized similarity matrix $QK^T$ encodes the geometric correspondences between $\tilde{H}_d$ and $\tilde{H}^{1:M}_r$. 
To enhance the representation, learnable positional embeddings are incorporated into $Q$ and $V$. The implicit warped texture features $I_w$ are then obtained by weighting $\tilde{I}^{1:M}_r$ with the learned geometric correspondences. Finally, a downsampling convolutional layer is applied to refine the features, formulated as:
\begin{equation}
I_{w}=Conv(reshape(Attention(Q,K,V)))
\end{equation}

\subsection{Confidence-Guided Fusion \& Generation}
In this stage, we propose a confidence-guided fusion (CGF) module that dynamically evaluates the relevance of reference-guided features from RAC and selectively integrates them with the warped results of DOFW. As illustrated in \Cref{pipeline} (c), the output of CFG is derived by fusing $E_w$ and $I_w$ accoding to the learnable fusion mask $M$, as follows:
\begin{equation}
F_w = M \otimes E_w+(1-M)\otimes I_w
\end{equation}
where $\otimes$ denotes the Hadamard product. The animated image $\hat{I}_d$ is then obtained by feeding the fusion output $F_w$ into the generator:
\begin{equation}
\hat{I}_d = \mathcal{G}(F_w)
\end{equation}
Following \cite{guo2024liveportrait}, we adopt SPADE decoder \cite{park2019semantic} as the generator $\mathcal{G}$, which is well-suited for generating high-quality images conditioned on semantic layouts. 

\subsection{Training}
\noindent \textbf{Training Strategy}. Since the DOFW directly deforms $I_s$ using estimated motion fields, it converges faster than the RAC. To address the resulting optimization imbalance, we adopt a progressive training strategy: the RAC is first trained alone to stabilize its attention mechanisms (Warm up phase), and then both modules are jointly optimized, ensuring balanced gradient updates and effective compensation learning (Joint adaptation phase).

\noindent \textbf{Optimization}.
Following previous methods~\cite{wang2021one,guo2024liveportrait}, our approach employs a combination of loss functions (perceptual loss $\mathcal{L}_P$~\cite{johnson2016perceptual}, adversarial loss $\mathcal{L}_G$~\cite{wang2019few,wang2018high}, and L1 reconstruction loss $\mathcal{L}_{rec}$) to enhance visual quality. The overall loss function is formulated as:
\begin{equation}
\mathcal{L} = \lambda_P\mathcal{L}_p +\lambda_G\mathcal{L}_G+\lambda_{rec}\mathcal{L}_{rec}
\end{equation}
where $\lambda_P$, $\lambda_G$, and $\lambda_{rec}$ are hyper-parameters balancing the contributions of each loss. 

\begin{table*}[htbp]
\centering
\caption{\textbf{Quantitative comparisons of self-reenactment}. R=1 indicates that a single reference frame is used during inference, and so on.}
\small
\resizebox{1\linewidth}{!}{
\begin{tabular}{*{11}{c}} 
\toprule
\multirow{2}{*}{Method} & \multicolumn{5}{c}{\textbf{VFHQ}} & \multicolumn{5}{c}{\textbf{HDTF}} \\
\cmidrule(lr){2-6} \cmidrule(lr){7-11} 
& LPIPS$\downarrow$ & PSNR$\uparrow$ & SSIM$\uparrow$ & L1$\downarrow$& FID$\downarrow$ & LPIPS$\downarrow$ & PSNR$\uparrow$  & SSIM$\uparrow$  & L1$\downarrow$& FID$\downarrow$ \\
\midrule
FOMM~\cite{wang2021one} & 0.5123 & 21.6110 & 0.7021 & 0.0490 & 156.9245 & 0.3342 & 24.8712 & 0.7865 &0.0334 & 111.5943 \\ 
Face Vid2vid~\cite{wang2021one} &0.4903  & 21.7473 & 0.7132 & 0.0487 & 126.0494 & 0.2771 & 26.4803& 0.8374 & 0.0294 & 71.063 \\ 
MCNet~\cite{hong2023implicit}  & 0.4498 & 23.1351 & 0.7530 & 0.0411 & 121.1011 & 0.2494 & 28.7923 & 0.8742 & 0.0215 & 79.9111 \\ 
X-Portrait~\cite{xie2024x} & 0.4298 & 23.1803 & 0.7597 & 0.0408 & 92.9037 & 0.2118 & 28.9298 & 0.8631 & 0.0284 & 63.8333 \\
LivePortrait~\cite{guo2024liveportrait} & \underline{0.3953} & 23.2907 & \underline{0.7662} & 0.0398 & \underline{31.3928} & \underline{0.1817} & \underline{29.1516} & \underline{0.8954} & \underline{0.0213} & \underline{36.4944} \\
AppMotionComp~\cite{zhao2025synergizing} & 0.4101 & \underline{23.4723} & 0.7566 & \underline{0.0379} & 82.8032 & 0.2677 & 28.2386 & 0.8654 & 0.0223 & 45.8095 \\
\midrule
Ours (R=1) & \textbf{0.2798} &  \textbf{24.7931} & \textbf{0.8207} & \textbf{0.0366} & \textbf{27.4209} & \textbf{0.1527} & \textbf{30.6826} & \textbf{0.9205} & \textbf{0.0203} & \textbf{34.7572}\\
Ours (R=2) & \textbf{0.2429} &  \textbf{25.4358} & \textbf{0.8396} & \textbf{0.0342} & \textbf{21.5998} & \textbf{0.1430} & \textbf{30.9842} & \textbf{0.9255} & \textbf{0.0197} & \textbf{32.3417}\\
\bottomrule 
\end{tabular}
}
\label{tab:self-quantitative}
\vspace{-12pt}
\end{table*}

\section{Experiments}
\label{sec:exp}
\subsection{Implementation Details}
\noindent \textbf{Datasets}. We conduct experiments on VFHQ~\cite{wang2022vfhq} and HDTF~\cite{zhang2021flow} datasets. Our model is trained on the VFHQ training set and evaluated on its test split. To further assess its generalization ability, we also report results on a curated subset of 50 videos from HDTF.

\noindent \textbf{Training Details}. We train our model for 150 epochs using the Adam optimizer~\cite{kingma2014adam} with a learning rate of $2\times 10^{-4}$, $\beta_1=0.5$, and $\beta_2=0.999$. Additionally, the variance $\sigma^2$ in \Cref{gaussian} is assigned to 0.01.

\noindent \textbf{Evaluation Metrics}. Following \cite{guo2024liveportrait}, we adopt Structural Similarity Index (SSIM)~\cite{wang2004image}, Peak Signal-to-Noise Ratio (PSNR), Learned Perceptual Image Patch Similarity (LPIPS), L1 distance and FID~\cite{heusel2017gans} to evaluate the generalization quality and motion transfer accuracy of portrait animation results.

\subsection{Comparison with State-of-the-art Methods}
\noindent \textbf{Baselines}. We compare our model with several state-of-the-art video-driven portrait animation models, including FOMM~\cite{siarohin2019first}, Face Vid2vid~\cite{wang2021one}, MCNet~\cite{hong2023implicit}, X-Portrait~\cite{xie2024x}, LivePortrait~\cite{guo2024liveportrait} and AppMotionComp~\cite{zhao2025synergizing}. We assess performance in both self- and cross-reenactment at a resolution of $256\times256$, with comparisons to animated portraits and ground truth images detailed qualitatively and quantitatively.

\begin{figure}[htbp]
\centering
\includegraphics[width=1\linewidth]{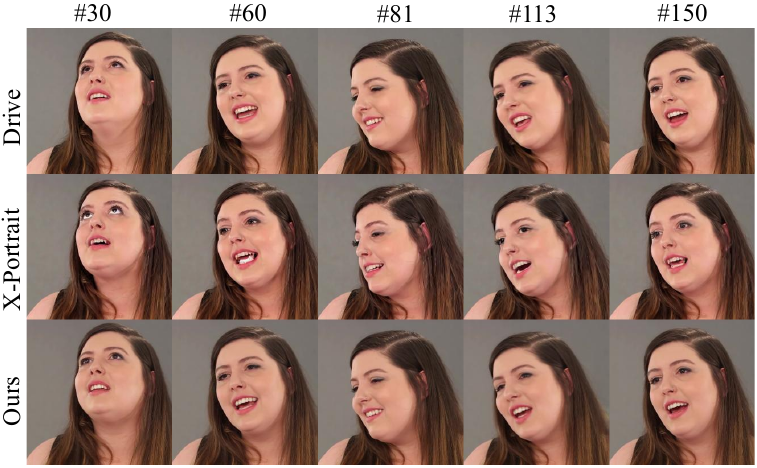}
\caption{\textbf{Temporal consistency evaluation}.}
\label{time-fig}
\vspace{-10pt}
\end{figure}

\noindent \textbf{Self-reenactment}. For self-reenactment evaluation, we employ the first frame of each test video as the source image and generate the complete video sequence, while each subsequent frame simultaneously serves as both the driving image and the ground truth for quantitative evaluation. \Cref{tab:self-quantitative} shows the quantitative comparisons, our method numerically surpasses the other methods on all metrics on VFHQ and HDTF. As depicted in \Cref{qualitative} (a), our qualitative comparisons highlight the strengths of our approach, our approach leverages multiple frames to fill in missing parts of the source image, and the RAC effectively corrects backgrounds distorted by the explicit branch. The examples show our model’s ability to transfer motion details like gaze direction (row1 and row4) and lip movements (row 3), while also supplementing missing parts (row 2). Moreover, our approach achieves superior temporal consistency (as shown in \Cref{time-fig}), further validating its overall effectiveness in generating stable and coherent facial animations across diverse sequences.

\noindent \textbf{Cross-reenactment}. We curate a cross-reenactment test set by randomly selecting 50 pairs from the VFHQ and HDTF test sets, ensuring that the source and driving images originate from different individuals. \Cref{qualitative} (b) shows the qualitative comparisons of cross-reenactment. Diffusion-based X-Portrait~\cite{xie2024x} often produces exaggerated expressions due to the lack of explicit motion modeling, while AppMotionComp~\cite{zhao2025synergizing} struggles to accurately mimic the driving image when the required appearance or motion patterns are absent from its codebooks.
Benefiting from our proposed hybrid framework, our method demonstrates more precise facial motion control (row 1 and row 2) and effectively addresses the issue of low-fidelity facial animation caused by head rotations (row 3 and row 4).

\subsection{Ablation Study}
In this section, We evaluate the 3D implicit keypoints correction and the fusion mechanism on the VFHQ and HDTF test sets. As reported in \Cref{tab:ablation-kp}, replacing 3D implicit keypoints with 2D sparse keypoints detected by FOMM~\cite{siarohin2019first} markedly worsens LPIPS, PSNR, SSIM, and FID, while increasing the reference frames from R=1 to R=2 further improves all metrics, demonstrating the advantage of 3D geometry and multi-reference input for robust spatial alignment. For fusion mechnism, \Cref{tab:ablation-fusion} compares our method with channel concatenation, simple summation, and masked summation on the HDTF test set, our approach achieves the best scores across all measures, confirming the effectiveness of the proposed fusion strategy for high-fidelity reenactment.

\begin{table}
  \centering
  \caption{Ablation study for the dimension of keypoints.}
  \small
\resizebox{1\linewidth}{!}{
  \begin{tabular}{@{}lcccccc@{}}
    \toprule
    Module & LPIPS$\downarrow$ & PSNR$\uparrow$ & SSIM$\uparrow$ & L1$\downarrow$& FID$\downarrow$\\
    \midrule
    2D-based (R=1) & 0.6569 & 21.2332 & 0.6868 & 0.0514 & 246.6325  \\
    3D-based (R=1) & \underline{0.6004} & \underline{21.4362} & \underline{0.7112} & \underline{0.0497} &  \underline{198.8629} \\
    3D-based (R=2) & \textbf{0.4489} & \textbf{23.6507} & \textbf{0.7844} & \textbf{0.0419} & \textbf{115.4916}\\
    \bottomrule
  \end{tabular}}
  \label{tab:ablation-kp}
  \vspace{-10pt}
\end{table}

\begin{table}
  \centering
  \caption{Ablation study for the fusion mechanism.}
  \small
\resizebox{1\linewidth}{!}{
  \begin{tabular}{@{}lcccccc@{}}
    \toprule
    Method & LPIPS$\downarrow$ & PSNR$\uparrow$ & SSIM$\uparrow$ & L1$\downarrow$& FID$\downarrow$\\
    \midrule
    Concatenate & 0.1583 & 28.7014 & 0.9037 & 0.0233 & 39.8241  \\
    Sum & 0.1630 & \underline{29.9015} & 0.9104 & \underline{0.0213} & 45.4438 \\
    Sum-mask & \underline{0.1567} & 29.2962 & \underline{0.9156} & 0.0243 & \underline{35.0302} \\
    Ours &\textbf{0.1430} & \textbf{30.9842} & \textbf{0.9255} & \textbf{0.0197} & \textbf{32.3417} \\
    \bottomrule
  \end{tabular}}
  \label{tab:ablation-fusion}
  \vspace{-10pt}
\end{table}
\section{Conclusion}
\label{sec:conclusion}
In conclusion, we present SynergyWarpNet, a unified framework for video-driven portrait animation that cooperatively integrates explicit warping and attention-guided refinement. The DOFW ensures precise facial feature preservation during animation reproduction, while the RAC effectively compensates for missing details in source images through multi-reference image utilization, simultaneously addressing facial blurring artifacts caused by global warping operations. Otherwise, a confidence-guided fusion network is proposed to adaptively fuse the warped results of these two component. Comprehensive quantitative evaluations and qualitative assessments consistently validate the effectiveness of our method, showing significant improvements in both visual fidelity and temporal coherence. 


\vfill\pagebreak

\bibliographystyle{IEEEbib}
\bibliography{strings,refs}

\end{document}